\documentclass[runningheads]{llncs}
\usepackage{graphicx}
\usepackage{bibentry}
\usepackage[authoryear]{natbib}
\usepackage[hyphens]{url} 
\usepackage{amsmath}
\usepackage{eurosym}  
\usepackage{subfig} 
\usepackage{array}
\newcolumntype{P}[1]{>{\centering\arraybackslash}p{#1}} 
\usepackage{threeparttable}   
\usepackage{booktabs} 
\usepackage{arydshln} 
\usepackage{xcolor} 

\nobibliography*

\begin{document}
\title{Timing Process Interventions with Causal Inference and Reinforcement Learning \\ 
}
\titlerunning{Timing Process Interventions with CI and RL}

%
\author{Hans Weytjens\inst{1}\orcidID{0000-0003-4985-0367} \and
Wouter Verbeke\inst{2,3}\orcidID{0000-0002-8438-0535} \and
Jochen De Weerdt\inst{3}\orcidID{0000-0001-6151-0504}}

\authorrunning{H. Weytjens et al.}

\institute{Research Centre for Information Systems Engineering (LIRIS),\\ Faculty of Economics and Business,\\ KU Leuven, Leuven, Belgium\\
\email{\{hans.weytjens,wouter.verbeke,jochen.deweerdt\}@kuleuven.be}}

\maketitle

\begin{abstract} 
The shift from the understanding and prediction of processes to their optimization offers great benefits to businesses and other organizations. Precisely timed process interventions are the cornerstones of effective optimization. Prescriptive process monitoring (PresPM) is the sub-field of process mining that concentrates on process optimization. The emerging PresPM literature identifies state-of-the-art methods, causal inference (CI) and reinforcement learning (RL), without presenting a quantitative comparison. Most experiments are carried out using historical data, causing problems with the accuracy of the methods' evaluations and preempting online RL. Our contribution consists of experiments on timed process interventions with synthetic data that renders genuine online RL and the comparison to CI possible, and allows for an accurate evaluation of the results. Our experiments reveal that RL's policies outperform those from CI and are more robust at the same time. Indeed, the RL policies approach perfect policies. Unlike CI, the unaltered online RL approach can be applied to other, more generic PresPM problems such as next best activity recommendations. Nonetheless, CI has its merits in settings where online learning is not an option.

\end{abstract}

\section{Introduction}
Moving from predicting the outcome of a running process to optimizing it with respect to a goal implies making decisions about actions that will change its course. In its most basic form, the optimization of a process assumes a correctly timed intervention (or sometimes non-intervention) in it. Examples range from the escalation of a customer complaint process to higher management echelons, the maintenance of a machine, a customer call to speed up an administrative process or to maximize turnover, an additional test to conduct to reduce a patient's length of stay at hospitals, etc. \textit{Prescriptive Process Monitoring} (PresPM) is a young subfield of Process Mining (PM) studying business process optimization methods. Optimization in the PresPM context concerns decisions an agent has to take to optimize the outcome of a running case given certain goals (metrics). It does not concern enhancing the underlying process itself, as practiced in PM. Our PresPM (see below) literature review reveals that two methods, \textit{reinforcement learning} (RL) \citep{PPM:RL} and \textit{causal inference} (CI) \citep{PPM:judea,PPM:imbens}, emerge as pathways. However, a quantitative comparison is currently missing. Most of the research on PresPM works with offline historical data creating two limitations: Online RL is not possible and experimental results prove difficult to quantify accurately for lack of counterfactuals.

This research gap defines our contribution. In our experiments, we introduce online RL to business processes and benchmark it against CI. Our use of synthetic data, rather than historical event logs as in earlier PresPM research, is not only instrumental in permitting both online RL and CI, but also enables deeper insights, a correct evaluation of the experiments' results, and the calculation of perfect policies as an absolute benchmark.

Solutions to \textit{timed process interventions} can be seen as a gateway to solving the more generic problem of recommending the next best activities in a process. In timed interventions, the agent has one chance to make an intervention sometime during the process, whereas, in next best activity problems, the agent has to choose between all possible activities at every step in the process. Both problem types are structurally the same, the RL algorithms for the former can be transferred to the latter problem without modification. The relative simplicity of timed process interventions in terms of combinatorial possibilities (state space) results in fewer data and computational requirements. It will permit easier insights into the characteristics of the used models and find faster real-world adoption. Furthermore, a vast number of relevant applications for timed process interventions exist. For these reasons, our experiments focus on timed process interventions, rather than next best activity recommendations.

This paper is structured as follows. Section~\ref{PPM:sec:literat} introduces concepts pertaining to PresPM and refers to related work around them. We then move on to the experimental Section~\ref{PPM:sec:exp} comparing RL to CI using two synthetic datasets. The insights gained from the literature review and experiments lead to a deeper discussion of the two PresPM methods in Section~\ref{PPM:sec:discussion}. We conclude this paper and suggest avenues for future work in Section~\ref{PPM:sec:concl}.

\section{Background and related work}\label{PPM:sec:literat}

\subsection{Preliminaries}\label{PPM:subsec:prelimn}

The goal of PresPM \citep{PPM:costaware,PPM:quovadis,PPM:when,PPM:nextbest} is to recommend actions for ongoing cases in order to optimize their outcomes as measured by a certain metric. Before the appearance of ML, most prescriptive problems relating to processes concerned industrial processes and were approached by operations research methods, requiring mathematical models \citep{ppm:ORbook} of the problems. Later, ML opened new, model-free opportunities, leading to a substantial body of research about its application in predictive maintenance and process control (regulating a system to keep certain parameters within a defined range). For overviews, the interested reader is referred to \cite{PPM:CARVAL} and \cite{PPM:four}. Within PresPM, a much more recent discipline, CI and RL emerged as two promising methods for process outcome optimization and will be the subject of this research. 

The business processes studied in PresPM exhibit a substantial degree of variation, in practice often explained by the humans in the loop. We believe that the high degree of variation in these processes makes them generic, and hence, the results of this work also apply to more constrained or structured processes such as industrial processes.

In this work, we restrict our focus to the optimization of a single process in isolation. This is an important assumption given that in practice, many processes affect and even interact with each other. Action recommendations may impact each other e.g., in case of limited resources. To the best of our knowledge, the very large majority of PresPM research assumes process independence. We also assume that decision points in processes are known after either consulting experts and/or applying PM techniques to identifying decision rules \citep{conc:decmin} and causal relationships (e.g., \cite{PPM:root}, \cite{PPM:reason}, \cite{intro:meaning}, \cite{intro:poly}.

\subsection{Causal inference}\label{PPM:subsection:CI_lit}

By default, CI \citep{PPM:judea,PPM:imbens} works with offline, logged data. The field can be subdivided into two components. The first concerns the detection of causal relationships: ``Which treatment(s) have an effect on the process' outcomes?''. The second CI component involves estimating the effect of treatments. We concentrate on the \textit{individual treatment effect} (ITE) \citep{PPM:ITE}, which is the difference between predicted outcomes of (possible) treatment(s) and non-treatment for a given sample. For example, when our model predicts that calling (treatment) customer $x$ will increase revenue by $200$\euro, while $x$ is expected  to reduce sales by $100$\euro{} if not called (non-treatment), then the ITE$_x$ is $300$\euro{}. Note that the ITE is an expectation, not a hard-coded causality. Usually, a threshold (e.g., $50$\euro{} in our example) is determined to arrive at a policy for selecting (non\nobreakdash-) treatments. The main challenge is the absence of \textit{counterfactuals} in the dataset. A counterfactual is the unobserved outcome of a case assuming another treatment than the one factually applied (not to be confused with negative or forbidden events or traces as in \cite{PPM:neg}). In the absence of \textit{randomized controlled trials} (RCTs), realized by a policy of random interventions, \textit{selection bias} will occur as the data-gathering policy leads to different distributions of treatment and non-treatment samples in the datasets. Combating selection bias is an important aspect of CI (e.g., \cite{PPM:ITE}). Real-world CI applications include marketing (e.g., churn reduction:~\cite{PPM:Wouter}, discounting, ...), education (e.g., \cite{ppm:edu}), recommender systems, etc. Most of these applications, however, are cross-sectional rather than longitudinal: There are no timing issues, let alone sequential treatments as seen in processes.

To the best of our knowledge, CI plays no role in the optimization of industrial processes. In line with the overview provided by \cite{PPM:quovadis}, we found no papers in the sparse PresPM literature published before 2020 claiming to use CI for process outcome optimization. \cite{PPM:recommend} and \cite{PPM:nextbest} did apply a form of \textit{indirect CI} with that aim, albeit without carrying the CI label. The indirect approach consists of first predicting the most likely (or distribution of) suffix(es) for every possible treatment given a certain prefix. In the second step, another model predicts outcomes for all these suffixes, which will then be used to choose a treatment. In the \textit{direct} CI approach, the process outcomes for all possible treatments for a given prefix are directly predicted. Direct CI implementations can be found in \cite{PPM:ppmci} (without timing considerations) and \cite{PPM:costaware} and \cite{PPM:when} (including timing). With the exception of \cite{PPM:ppmci}, none of the PresPM papers addresses selection bias. \cite{PPM:seq2seq}, in contrast, use a sequence-to-sequence recurrent NN that automatically builds a treatment-invariant representation of the prefixes to combat the selection bias in a medical treatment problem.

The lack of counterfactuals in the test set stemming from the use of offline data hinders the accurate evaluation of CI methods' results: For a given prefix, the action recommended by the CI model may be absent from the cases in the test set. Researchers cope with this problem by relying on a predictive model to estimate outcomes, a distance-minimizing algorithm to find the nearest case in the training or dataset, or a generative model that produces augmented data \citep{PPM:Real}.

\subsection{Reinforcement learning}\label{PPM:subsection:RL_lit} 

RL \citep{PPM:RL} is an important class of ML algorithms learning policies that guide an agent's behavior or sequence of actions in an environment in order to maximize an expected cumulative \textit{reward}. Early successes in computer games drew much attention to RL, which has since then expanded not only into industrial processes but also into many other fields such as robotics (e.g., autonomous driving:~\cite{ppm:RLcar}), healthcare \citep{ppm:med}, engineering, finance, etc. RL comes in many flavors. We will discuss and use the widespread \textit{Q-learning} variant. In processes, the most important reward is often the process outcome that becomes known at the conclusion (last event) of the case. Regardless, intermediate rewards could be easily included in RL should they occur. The cost of actions can be viewed as a negative reward. At its core, RL assumes an \textit{online} environment that the agent can interact with. RL does not need an environment ($\rightarrow$ process) model. Instead, real (or simulated) \textit{episodes} ($\rightarrow$ cases) are just executed and their rewards ($\rightarrow$ outcomes) are observed. For every encountered \textit{state} ($\rightarrow$ prefix), a \textit{state-action value} (Q) is learned for every possible \textit{action}. Q represents the state-action value for the next state ($\rightarrow$ prefix) plus the reward minus the cost of that action to get to that next state (a \textit{transition}). For any given prefix, the state-action values can be interpreted similarly as the effects of the possible treatments learned by CI. The difference between the state-action value for a treatment and the one for the non-treatment corresponds to the ITE at that state (prefix). The state-action value of the last prefix of a (completed) process is its final outcome. Given the size of the state space ($\rightarrow$ number of possible prefixes) in most processes, these state-action values cannot be stored in tabular form (\textit{Q-table}). Instead, they are approximated by an NN. This is called \textit{deep reinforcement learning} \citep{PPM:replay}. At every state ($\rightarrow$ prefix), the policy will be to choose the action with the highest relative state-action value. Learning is achieved by playing out many processes and iteratively updating the Q-table NN after each (batch of) observed rewards ($\rightarrow$ outcomes). In order to explore all areas of the state space and to prevent prematurely settling into a sub-optimal policy, a certain degree of exploration is introduced: The agent will sometimes overrule the policy and choose another action, especially at the beginning of the learning process. RL has found many applications in process outcome optimization, e.g., in robotics \citep{PPM:robotRL} and industrial process control \citep{PPM:PARA,PPM:Spielberg}, but few researchers \citep{PPM:learning,PPM:replace} apply RL to PresPM process optimization. 

In practice, the real-life form of data gathering is often too slow and too expensive. It can even be dangerous at the early stages of learning when the NN is insufficiently trained and significant exploration happens. An entire spectrum of alternative data-gathering methods at different proximities to reality exists. In academia, synthetic data are instrumental in investigating and comparing methods. For instance, \cite{PPM:PARA} work with synthetic data for industrial process control. Simulation models can be rooted in the laws of physics or even social sciences. \cite{PPM:Spielberg} use simulation in industrial processes, \cite{PPM:quadsim} use simulation to train robots and investigate pathways to close the \textit{reality gap}, the mismatch between the reality and the simulation. In \cite{PPM:liso}, a Fogg behavior model is used for healthy habit formation for cancer patients. Digital twins \citep{PPM:twins} are extended simulations benefiting from imputed real-time data. \cite{PPM:BURG} deploy a miniature factory in which their agent can act. In practice and academia alike, there is great interest in \textit{offline} RL i.e. to work with data from existing datasets (supervised data). This can be achieved by mining models from the data. PM discovery techniques, for example, yield grid graphs of business processes as representations of the agent's environment in \cite{PPM:learning} and \cite{PPM:replace}. Research on offline RL \citep{PPM:offline} also suggests using predictive models trained on the dataset to guide the agent through its environment and estimate outcomes, somehow similar to indirect CI. Alternatively, nearest-neighbors algorithms can force the agent to remain in the vicinity of the data-gathering policy. The RL agent can even be forced to remain within the boundaries of the dataset, which of course makes it harder to improve upon the original data-gathering policy. With the exception of purely synthetic data, these data-gathering strategies can all be used to derive a policy with which to initialize an online learning agent. This dual strategy greatly speeds up training, avoids expenses, and minimizes mistakes.

\subsection{Problem complexity}\label{PPM:subsection:complexity}
From the analysis of the related work, we identify two main drivers for problem complexity: \textit{action width} and \textit{action depth}. These drivers facilitate understanding problems at hand and will be helpful in positioning our experiments (Section~\ref{PPM:sec:exp}) and the subsequent discussion of CI and RL (Section~\ref{PPM:sec:discussion}).
Action width relates to the number of different actions (set size) available to an agent: Actions can be binary (\textit{interventions} e.g., ``apply'' or ``don't apply''), continuous (e.g., a regulating valve) or multi-class (e.g., several options such as ``visit'', ``call'' or ``email'' customer or ``do nothing''). In this paper, we define multi-class actions as \textit{treatments}. Interventions are thus a binary subclass of treatments. Asymptotically, the set of multi-class actions becomes the set of all possible activities, even including attributes, as in the next best activity prediction. The action depth is a measure of the longitudinal dimension and depends on how many consecutive actions can be taken within a process, and when. An action's timing can be predetermined (fixed or irrelevant) or an action can happen once at any time during the process' lifetime (one-off, as in timed process interventions). Sequences of (repeated) actions are another, more complicated, setting. Finally, actions can be continuous such as steering an autonomous vehicle to keep it in its lane.

\subsection{Research gap}\label{PPM:subsection:gap}
There exists no quantitative comparative analysis of CI and RL process outcome optimizations (nor any other) problems. This is the main research gap we address in this paper. As explained in subsection~\ref{PPM:subsection:CI_lit}, the use of historical data for the test sets hampers the evaluation of CI methods for lack of counterfactuals.  A similar issue appears in the RL literature that exhibits a prevalence of non-real-life work. Here, simulations or models based on reality are used to train and test online models without considering the performance on the original problems, thus ignoring the reality gap (exception: \cite{PPM:Spielberg}). We also address this issue by making use of entirely artificial synthetic data in our experiments. This form of data allows us to accurately evaluate CI, to test online RL and eliminate the reality gap, and to share the same test set between both methods. Additionally, none of the aforementioned papers compared their results to perfect policy results needed to gain an intuition for the absolute performance of their methods. This can be explained by the majority of the discussed papers treating rather complex problems for which computing such a perfect policy is intractable, hence the need for techniques such as CI and RL. We opt for timed interventions, which have narrow action widths (binary) and shallow action depths (at most once per process) so that we can easily compute results for a perfect policy. The next Section~\ref{PPM:sec:exp} describes our contribution: making an accurately evaluated CI-RL-perfect-solution comparison based on synthetic datasets.


\section{Experimental comparison of CI and RL}\label{PPM:sec:exp}


In the following three subsections, we describe our data generation, experimental setup, and results. 

\subsection{Data generation}
We work with synthetic processes generating the environment and data for our experiments in order to compute counterfactuals that are not available in real-world data. Knowing the counterfactuals allows for accurate evaluations of the experiments. Moreover, given a sufficiently small state space, a perfect policy can be derived and used to judge the absolute performance of methods. The same synthetic generative model can create both the offline dataset for CI and the online environment required for online RL. We first describe the two processes and then motivate our choice.
\subsubsection{Two synthetic processes}

\begin{figure}%
    \centering
    \subfloat[\centering Process model representing \texttt{Process\_1}]{{\includegraphics[width=7cm]{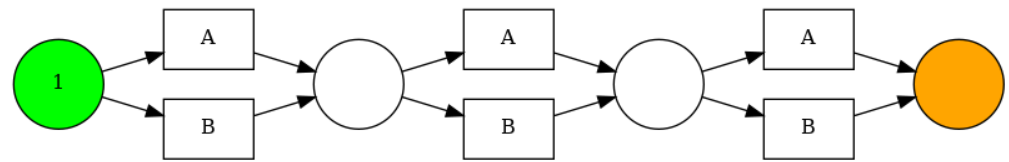} }}%
    \vspace{.5cm}
    \subfloat[\centering Process model representing \texttt{Process\_2} ]{{\includegraphics[width=9.5cm]{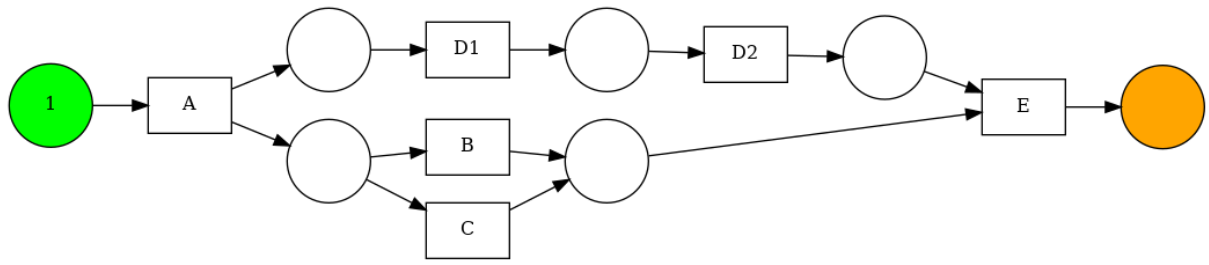} }}%
    \caption{Synthetic generative processes: Petri net visualization}%
    \label{PPM:fig:graphs}%
\end{figure}

The process models as Petri nets and key features of our two synthetic processes are shown in Figure~\ref{PPM:fig:graphs} and Table~\ref{PPM:tab:data} respectively. \texttt{Process\_1} is a sequence of three activities, either ``$A$'' or ``$B$'' with an according integer attribute. At one of the three events, a (free) intervention can be made. The outcome of the process is the sum of the attributes, where the attribute of the event where the intervention took place is multiplied by $2$ if activity ``$A$'' occurred at least once in the process, otherwise by $-2$. \texttt{Process\_2} consists of five events and includes both an AND and an XOR construct. Every \texttt{Process\_2} case carries an integer case attribute known from the start. Event attributes are integers as well, and an intervention can be made once in a process at any event. When an intervention is made (at a cost of $5$), the attribute corresponding to the first of ``$D1$'' or ``$D2$'' to occur thereafter (if any) will be multiplied by $2$ in case the process passes through its ``$B$'' branch, by $-4$ otherwise. The final outcome is the sum of the attributes times the case attribute.

\begin{table}[ht!]
     \begin{center}
     \caption{Synthetic generative processes: key features}
     \scalebox{0.8}{
     \begin{tabular}{ p{.26\textwidth}>{\centering}p{.31\textwidth}>{\centering\arraybackslash}p{.40\textwidth} }
     \toprule
      & \textbf{\texttt{Process\_1}} & \textbf{\texttt{Process\_2}} \\ 
     \midrule
     \textbf{Structure}&sequence of OR&AND with OR in lower branch\\
     \midrule
     \textbf{Activities}&every event activity from $[A, B]$ with prob $[.25, .75]$& as per graph. ``$B$'' and ``$C$'' with prob $[.2, .8]$\\
     \midrule
     \textbf{Attributes} ($Att$) & uniformly random from $[0, 1, 2, 5]$&uniformly random from $[1, 2, 3]$ for ``$D1$'' and from $[1, 2, 3, 4]$ for ``$D2$'', else $0$\\
     \midrule
     \textbf{Case variables} ($CV$) &-&uniformly random integer from $[1-10]$
\\
     \midrule
     \textbf{State space size}&512&720\\
     \midrule
     \textbf{Interventions}&at all steps or not at all&at all steps or not at all\\
     \midrule
     \textbf{Intervention cost}&0&5\\
     \midrule
     \textbf{Intervention effect} (intervention at event $n$)&Multiply $Att_n$ by 2 in case an ``$A$'' occurred in the process, otherwise by -2.&$Att$ corresponding to next of ``$D1$'' or ``$D2$'' multiplied by $2$ (in case of ``$B$''-process) or $-4$ (in case of ``$C$''-process)
\\
     \midrule
     \textbf{Outcome} & $\sum Att$
     & $CV * \sum Att$\\
     \bottomrule
      \end{tabular}
      
      \label{PPM:tab:data}
      }
      \end{center}
      \end{table}

\subsubsection{Motivation}\label{PPM:subsec:mot}
\begin{figure}[ht]%
    \centering
    \includegraphics[width=7cm]{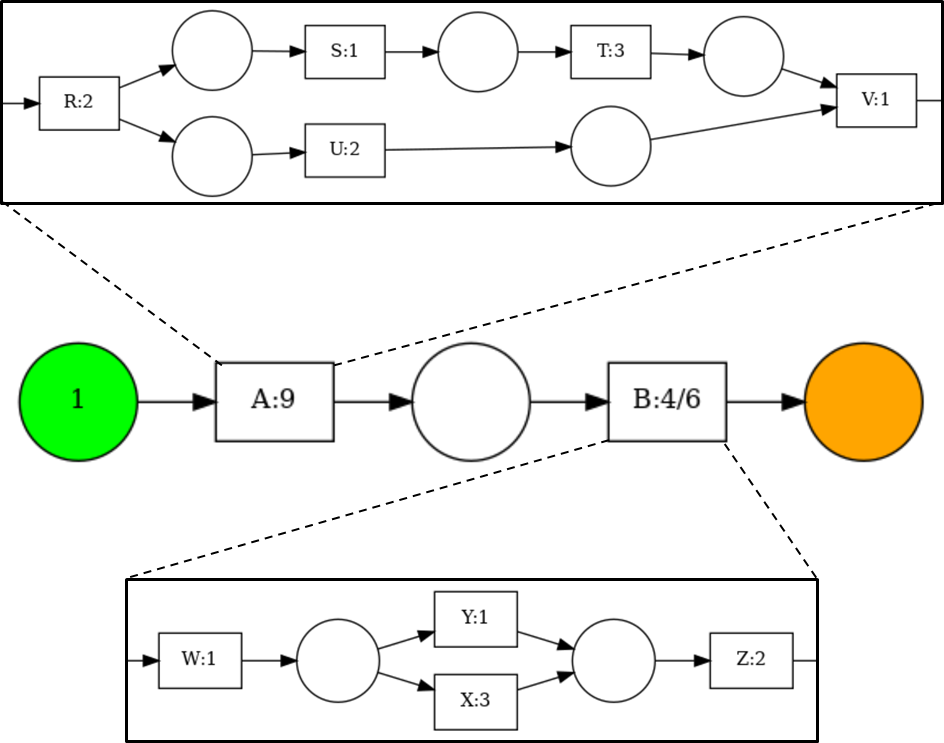} %
    \caption{Example of how a process can be reduced to a simpler process for PresPM purposes. ``$A$'' summarizes the subprocess without decision points ``$R-V$'', whereas ``$B$'' summarizes the subprocess `$W-Z$'' which includes a decision point ``$W$''.}
    \label{PPM:fig:reduce}%
\end{figure}

These processes and interventions were designed to be simple for clear insights, yet representative of real-world processes by incorporating their main challenges. Since PresPM concerns actionable decisions, we can reduce sub-processes that do not contain any decision points (and are not a branch of a parallel structure with another branch containing such a decision point) into one event (e.g., the subprocess ``$R-V$'' collapses into event ``$A$'' in Fig.~\ref{PPM:fig:reduce}), thus significantly shrinking the process model.  In our experiments, the interventions only change event attributes but in reality, they may alter the control-flow as well. That would not change the CI and RL algorithms. Moreover, when all control-flow variations starting from a given decision point (after event ``$W$'' in Fig.~\ref{PPM:fig:reduce}) merge together in one location/activity later in the process model (``$Z$'') without containing any further intermediate decision points, then they can be reduced to one event (``$B$'') as well. The value of this event's attribute will vary according to which decision was made and which control-flow variant was followed earlier.


Both processes have a strong stochastic component to reflect the uncertainty accompanying real-life processes.  The values of the three activities and attributes in \texttt{Process\_1} are sampled from probability distributions, whereas activities in \texttt{Process\_2} are governed by the given structure, with the attributes and case variables sampled from probability distributions as well. A real-life decision-maker is not only confronted by stochasticity, but the information available to make decisions may also differ between cases. Our synthetic processes also incorporate this aspect:  as long as no ``$A$'' appears in \texttt{Process\_1} or \texttt{Process\_2} hasn't passed through its ``$B$'' or ``$C$'' branch, it cannot be known for sure whether intervening will be beneficial or detrimental. 

As in many real-world processes, the outcomes of both processes will only be known at their conclusion. Including intermediary rewards or penalties, however, would not significantly alter the CI or RL algorithms. 

Our experiments investigate binary actions (interventions). This simplification allows for clearer insights without loss of generalization. As direct CI is generally not suited for sequences of actions, we further simplified by opting for one-off actions (timed interventions) to permit a CI-RL comparison; the RL method, however, can be extended to sequential or continuous actions without modification. In combination with the use of synthetic data, the small state space resulting from a narrow action width and shallow action depth also renders calculating perfect policies practical.

\subsection{Experimental setup}\label{PPM:subsec:exp}
\begin{figure}[!ht]
\center
\includegraphics[scale=.25]{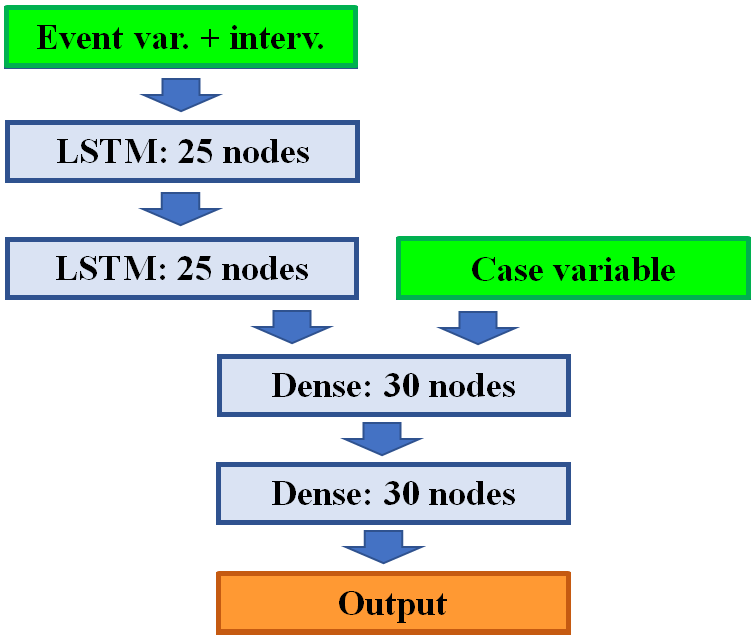}
\caption{The NN architectures (regression model) are almost identical for CI and RL. The first layer includes an additional ``intervention'' feature for CI. The last layer outputs a scalar (outcome) for CI and a 2-dimensional vector (Q for ``intervention´´ and ``non-intervention'') for RL.} \label{PPM:fig:NN}
\end{figure}

Even though increasingly performant next event prediction algorithms exist (e.g., \cite{PPM:cama}), the indirect CI approach inevitably compounds the errors of two successive prediction models. The direct CI approach circumvents the next event/suffix prediction stage. The simplicity of working with one model favors the direct approach and we used it in our experiments. We use one NN to predict process outcomes for CI; the intervention (Boolean) is part of its inputs, and the batch size is 1,024. RL is achieved with a standard Q-learning architecture with a 1,024-transition samples memory for stabilization (experience replay: \cite{PPM:replay}). Transition samples are retained in the memory according to the first-in-first-out (FIFO) principle. A penalty of 100 is applied for intervening more than once. An NN predicts Q for both possible actions (``intervention'' and ``non-intervention'') at every encountered state (prefix). For a balanced comparison, the same NN architecture is used for both CI and RL. Our NNs have two LSTM and two dense layers as displayed in Fig~\ref{PPM:fig:NN}. 

\begin{table}[ht]
\centering
\caption{Experimental settings.}\label{PPM:tab:expsetup}
\scalebox{0.8}{
\begin{tabular}{ll}
\toprule
Setting & Value \\
\midrule
Input features & ``activity'', ``attribute'', \\
&and ``case variable'' (for \texttt{Process\_2})\\
Categorical features representation& one-hot-encoding\\
Sequence length (padding for shorter prefixes)& 3 and 5 (\texttt{Process\_1} and \texttt{Process\_2})\\
\midrule
Loss function & MAE\\
Optimizer&ADAM\\
Metric & Uplift\\
Nr. runs per experiment &  5 \\
Batch size & 1,024\\
Memory size (RL) & 1,024\\
\midrule
Nr. epochs & early stopping\\
Patience & 5\\
Test set size & 1,000 samples \\
Validation set size (\% of  training set) &  20\% (threshold calculation in CI)\\
\bottomrule 
\end{tabular}
}
\end{table}

For the CI learning phase, an RCT dataset of 10,000 samples is generated. This largely exceeds both processes' state space size and should, therefore, offset CI's offline handicap. For RL, data are generated on the fly. The test set consists of 1,000 samples for which all counterfactuals are computed (feasible thanks to the relatively simple processes and the binary one-off action design). The data generated by the synthetic processes are preprocessed as follows: The activity levels are one-hot encoded. The outcomes, attributes, and case variables (\texttt{Process\_2}) are standardized.  For CI, the intervention decision ($1$ or $0$) is concatenated with the other event features. For every case sample, we build a sequence (sequence length = total process length) for every prefix, using padding to complete the sequence for ongoing process instances. We thus arrive at a two-dimensional data structure that is fed into the models' input layer. For \texttt{Process\_2}, the case variable enters the models separately after the LSTM layers.

Every experiment is carried out five times and learning stopped using an early-stopping algorithm for both methods. A policy based on CI requires identifying the threshold, which we identify as the value that maximizes the ITE score on a 20\% validation set. \textit{Uplift} \citep{PPM:Wouter, PPM:Wouteruplift} is the metric to evaluate the results. It is the difference between the process outcomes of implementing the policy and not intervening at all, cumulated over the complete test set. The experimental settings are summarized in Table~\ref{PPM:tab:expsetup}.

\subsection{Results}
We summarized our experimental results in Table~\ref{PPM:tab:results}. RL clearly outperforms CI for both processes: The mean scores significantly higher. The standard deviations of the RL scores are much lower, making RL by far the more robust method. Most or all of this outperformance can be attributed to RL's innate superior ability to find the optimal policy (see Section~\ref{PPM:sec:discussion}). The fact that online RL permits exploring all parts of the state space plays virtually no role here, as the CI training sets in our experiments contain the complete state space as well. Were this not the case, the observed CI-RL divergence would certainly widen.

\begin{table}
\centering
\vspace*{+3mm}
\caption{Experimental results comparing CI, online RL, a perfect  and random policy for both processes. Online RL reaches the highest uplift but requires much more computational effort than CI.}\label{PPM:tab:results}
\scalebox{.8}{
\begin{threeparttable}

\begin{tabular}{lc:cc:ccc}
  \cline{3-7}
  \multicolumn{2}{l}{}&\multicolumn{2}{c:}{\textbf{Uplift}}&\multicolumn{3}{c:}{\textbf{Computational effort}}\\
    \cline{3-7}
  \multicolumn{2}{l}{}&\textbf{Mean}& \textbf{StDev.}&\textbf{Unit}&\textbf{Mean}&\textbf{StDev.}\\
  \toprule
  &CI &1,526&36.8&epochs&188&27.6\\
  \textbf{\texttt{Process\_1}}&RL&1,616&5.0&transitions&6,000&2,549\\
    &\scriptsize{\textit{perfect}}&\scriptsize{\textit{1,651}}&&&&\\
    &\scriptsize{\textit{RCT}}&\scriptsize{\textit{-515}}&&&&\\
  \midrule
    &CI&1,682&203.1&epochs&212&51.7\\
  \textbf{\texttt{Process\_2}}&RL&1,806&34.7&transitions&26,800&10,628\\
  &\scriptsize{\textit{perfect}}&\scriptsize{\textit{1,845}}&&&&\\
  &\scriptsize{\textit{RCT}}&\scriptsize{\textit{-50,336}}&&&&\\
  \bottomrule
\end{tabular}

\end{threeparttable}
}
\end{table}

The precise knowledge of the (stochastic) synthetic generative processes enables computing perfect policies. This is done by drawing the complete state space in tree form and then calculating the best policy (intervene/don't intervene) from the leaves (512 or 720 for our processes) back to the prefixes of length one, always assuming no intervention happened before. Table~\ref{PPM:tab:results} shows that RL comes to within 3\% of the perfect policy results for both processes (some stochasticity is normal). The CI policy constitutes a substantial improvement over the RCT data-gathering policy that originally created the dataset as well, albeit to a lesser extent than RL.

Having set both RL's memory and CI's batch size to 1,024, one optimization step of the NN involves the same number of samples for both methods. Since every RL transition (except for the first 1,023 ones) was followed by an NN optimization step, we can directly compare the number of RL transitions to the number of CI epochs. Table~\ref{PPM:tab:results} shows that RL's computational requirements are two orders of magnitude higher than those for CI.

\section{Discussion}\label{PPM:sec:discussion}

In this section, we discuss the suitability of CI and RL for PresPM and show why RL outperformed CI in our experiments. We also address the issues of RL's online requirement, reward specification, and inefficiency.

\subsubsection{Causal inference}\label{PPM:subsection:CI_dis}
\begin{figure}[ht]%
    \centering
    \subfloat[\centering CI: policy requires threshold (TH), e.g., TH=0 $\rightarrow$ intervene at 1st event; TH=2 $\rightarrow$ intervene at 2nd event.]{{\includegraphics[width=5.5cm]{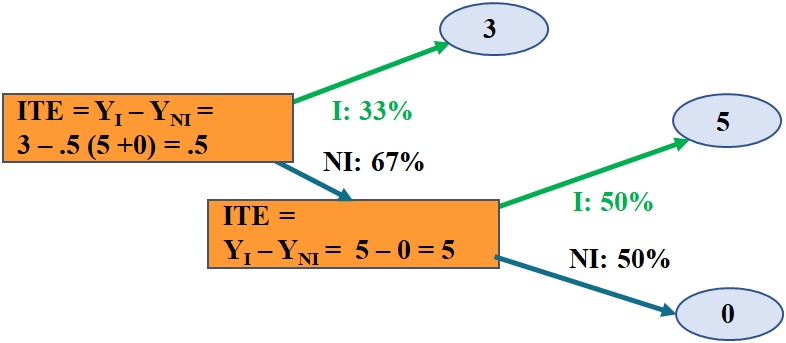} }\label{PPM:fig:CI}}
    \qquad
    \subfloat[\centering RL: optimal policy follows maximum Q values: do not intervene at 1st event, intervene at 2nd event.]{{\includegraphics[width=5.5cm]{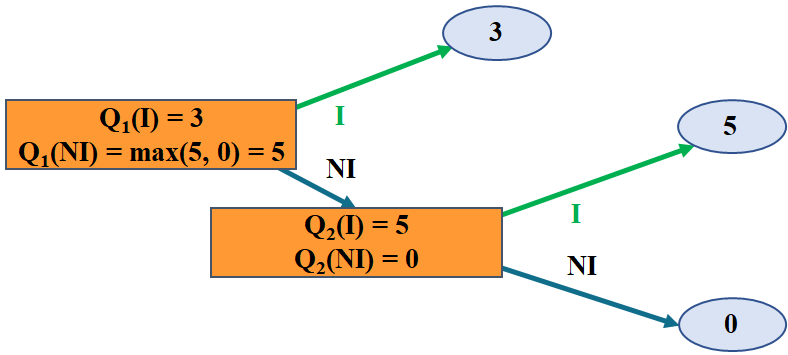} }\label{PPM:fig:RL}}%
    \caption{Simple process to compare CI to RL. Both agree on the policy at the second event. At the  first event, CI correctly estimates the outcome for intervention ($Y_{I}$), whereas the prediction model for the outcome for non-intervention ($Y_{NI}$), will observe two different outcomes ($5$ and $0$) and \textbf{summarize} (here: average) those to $2.5$. This value depends on the loss function and the samples' distribution (percentages in the graph) in the training set, which itself depends on the data-gathering policy. In contrast, RL selects the \textbf{maximum} of the two Q values in the second event.}
    \label{PPM:fig:simple}%
\end{figure}

Learning counterfactuals and treatment effects is at the core of CI. The sequential aspect of processes, however, poses a problem: The decision to not treat at a certain time in a running process does not preclude treatments later on in the process. For any given prefix in our experiments, direct CI relied on a predictive model to estimate the process outcomes for both intervention and non-intervention. This is problematic in the latter case: The predictive model cannot discern the optimal path from that prefix, and will instead consider the outcomes for all encountered treatments under the data-gathering policy that produced the relevant samples in the training set, as illustrated in the simplified example in Figure~\ref{PPM:fig:CI}. 

Direct CI, therefore, only operates safely on problems without any action depth (``fixed'' or ``irrelevant''), and will become increasingly suboptimal when moving to real processes with action depths ``once'' or ``multiple''. The action width for CI realistically comprises ``binary'' and ``multi-class'' treatments. CI cannot handle permanently-running processes. Thresholds are sub-optimal compromises and products of optimization algorithms themselves. Dependencies between processes, e.g., when resources (space, manpower) are limited or processes interact with each other, cannot be incorporated in the CI framework. Because of these deficits, optimal policies are theoretically out of CI's reach, as confirmed in our experiments. Nevertheless, CI policy results are still better than those that the data-gathering policy yields.

Similar to all other predictive models used for prescriptive or decision-making purposes, feedback loops \citep{PPM:hidden} risk deteriorating results: Implementing the CI policy will progressively shift the real-life data distribution away from the original training data, decaying the models' predictive accuracy. Frequent updates of the CI models would help but at the same time introduce new bias in the data (new data-gathering policy). However, with a sufficient degree of randomness in the decisions taken (as in RCTs and similar to exploration in RL), this iterative, in the limit \textit{online} CI, approach would neutralize the feedback loops.

\subsubsection{Reinforcement learning}\label{PPM:subsection:RL_dis}

 RL has many theoretical advantages over CI. It does not require a prediction model and can rely on observed outcomes. RL is entirely generic: Theoretically, it can deal with any action width or depth as well as with continuous processes. Next best activity prediction, which represents the ultimate action width and depth, requires no change to the RL algorithms we used for timed process interventions. RL models are very flexible: Constraints, rewards, and penalties can be added at liberty to avoid detrimental or unacceptable actions, pursue secondary goals, etc. With online RL, agents can freely interact with their environment, and dependencies between processes can be taken into account if the processes are treated concurrently by one model. Exploration in online RL theoretically visits the complete state-space (all possible prefixes). Given sufficient exploration, online RL policies will automatically adapt to a changing environment (concept drift). Proven theorems even show that online Q-learning algorithms converge given enough time. Both online and offline RL, however, are known to be inefficient, requiring many transitions to converge to the optimal policy, as demonstrated by our experiments. 
 
 The max operator over the Q-values (see Figure~\ref{PPM:fig:RL}) explains RL outperformance versus direct CI with equal data access. For every prefix, the learned Q values represent the expected outcomes for intervention and non-intervention respectively, assuming a (calculated) \textit{perfect} policy after that, whereas the ITEs in CI represent the difference between the expected outcomes, each of which depends on the sample distribution from the \textit{data-gathering} policy and the loss function. Note, however, that with an online CI approach (with real-time updating after every finished process observed) and allowing exploration, this data-gathering policy would converge to the optimal policy as well, thus practically obliterating the differences between CI and RL.

\subsubsection{Real-world implementation}\label{PPM:subsection:impl}
Despite its power and versatility, RL suffers from some important drawbacks. Yet, many of these are not entirely unique to RL but apply to CI and PresPM in general as well. The first such drawback is the risk of committing errors during real-time implementation. This implementation risk, however, can be reduced to that of the data-gathering policy (the de facto policy in place upon which the CI dataset is based) by inserting constraints into the RL algorithm that can easily deal with those. Rules mined earlier with a process discovery algorithm can frame the agent's actions as prescribed by \cite{PPM:manifesto}. Even human intuition can be inserted by allowing the human agent to overrule the RL algorithm's proposed action. In other words, implementing RL should not be riskier either than the original, existing policy or than implementing CI. The latter two policies occasionally make or propose costly mistakes too. If necessary, a two-stage offline-online approach can further reduce the risk: Offline RL based on simulations or predictive models can serve as an initialization to an online RL that then continues to learn acting in the real world, thereby closing the reality gap.

A similar argument can be made for the related challenge of reward specification. The desired outcome for a process to be optimized will not always be one-dimensional: The primary goal may be to reduce throughput time, however, without compromising employees' well-being and product quality. Moreover, such goals may shift over time or may need adjustment in the face of concept drift. Again, this challenge is not unique to RL, and exists regardless of the solution method, if any. When possible, these goals will be consolidated into one metric for use by both CI and RL. If not, RL can be extended to include constraints on undesired actions and/or rewards/penalties that promote secondary goals. As before, the human agent can also overrule the RL's algorithms suggestions.

RL is inefficient: It is data-hungry and slow to converge. Our experiments were based on relatively short and simple processes. Longer and more complicated processes (great action width/depth) will have an exponentially larger state space, suggesting that RL will no longer be a viable option where CI could still be. Yet, in deep RL, the Q-table is replaced by an NN, which to some extent obsoletes the need to visit the complete state space as unseen state-action (prefix-action) pairs can be interpolated. Working examples of this are video games with very large, and autonomous driving with near-infinite state spaces. The more similar regions the state space contains, the better this will work. Additionally, limiting the number of actions to the most relevant ones with causal discovery techniques (first CI component in \cite{PPM:judea}) may be a worthwhile investment before starting with RL (and CI as well). PM has an arsenal of causal discovery techniques that can be used to this end \citep{conc:decmin,PPM:root,PPM:reason}.

Online RL implies working with event streams rather than event logs. Streaming is an active field of PM research. \cite{intro:stream_confcheck} identifies different types of incomplete cases in the observation window. This is not an issue for an online RL as it always starts from a case's beginning and updates itself after observing rewards for every new event it encounters (transition) until the case is complete. 

The process independence assumption underlying both methods warrants caution when generalizing the results from our experiments. The larger the dependencies between processes and the larger the share of processes being optimized, the higher the risk of mutual process interference jeopardizing the expected results.

\section{Conclusions and future work}\label{PPM:sec:concl}
We conducted experiments on timed process interventions with synthetic data that render genuine online RL and the comparison to CI possible and allow for an accurate evaluation of the results. We showed how the theoretical problems burdening CI can be overcome by online RL, contingent upon the strong assumption of real-time implementation of the learned policies in the real world. In our experiments, online RL produced better and more robust policies than CI. In fact, RL nearly reached the theoretically optimal solution, which can be inferred because of the use of synthetic data. The used RL methods can also be applied without any modification to similar problems with greater action width and depth (next best activity prediction in the limit). When computational effort and/or the real-time implementation requirement preclude online RL, CI may be a viable alternative in scenarios where the dataset covers a large and evenly distributed share of the state space and action depth is limited.

With this work, we contributed to the nascent field of PresPM. We chose a simplified setting to gain some important insights. Reaching PresPM maturity will depend on exploring other, perhaps more sophisticated approaches, in ever more realistic settings. Further extensions of this work are, therefore, plentiful. First, an initial investigation of the merits of loss attenuation \citep{PPM:unc}, uncertainty \citep{PPM:unc}, and future individual intervention effects \citep{PPM:when} revealed promising insights but should be corroborated. Future work could also shed light on the conditions under which RL remains efficient enough on realistic problems with sequences of multiple possible actions (greater action width and depth). Further complications could include the introduction of outcome noise, uncertain inputs, and concept drift. Since the rewards of processes often only happen (or become known) at their conclusion, MC learning (as in \cite{PPM:learning}) could be a faster alternative to the classical Q-learning we used. The FIFO principle for the online RL transition samples memory could be replaced by more sophisticated sampling techniques such as described in \cite{concl:logsample} for PredPM, or in \cite{PPM:effects} for experience replay in RL. Leveraging uncertainty estimates could be another option to improve sampling. RL does adapt to concept drift, but only very slowly. As a consequence, RL is not suited to deal with disruptions (e.g., caused by a pandemic). Digital twins for processes or organizations have been proposed as a solution \citep{conc:digtwin} and are an avenue for future research. Instead of including the complete state space in the data for CI, as we did, it could be investigated to what extent CI would fall further behind online RL when the dataset only covers part of the state space (and contains selection bias caused by the data-gathering policy). For applications where online RL is not an option, more research on offline RL is recommended. Lifting the assumption of process independence would move the problem setting even closer to reality and would pose additional challenges: Process independence is a requirement to satisfy the \textit{stable unit treatment value assumption} (SUTVA) \citep{PPM:imbens} in CI. The combinatorial explosion caused by interdependent processes is challenging for RL as well and possibly demands additional heuristics (e.g., \cite{PPM:huang}). In the domain of CI, adaptations to the standard algorithms could lead to more capabilities in terms of action depth (possibly with a discounting mechanism as used in RL). Indirect CI's theoretical ability to handle sequences of actions could be weighed against the accuracy loss due to the compounding of two predictive models.  Combating selection bias in processes (as in \cite{PPM:seq2seq} for an environment without exogenous actors) beckons more research as well. Causal Reinforcement Learning \citep{PPM:CRL} enriches RL with the first component of CI (causal relationship detection) by means of causal graphs. It requires either a priori causal graphs (which are rarely available in PresPM, or deriving them from the observational data under a set of assumptions (e.g. \cite{PPM:discovercausal} and \cite{PPM:meet} for business processes). In our discussion about action width and depth (Subsection~\ref{PPM:subsection:complexity}), we did not elaborate on how the decision points and the set of possible actions available to the agents at those points are elaborated. Next to human expertise, both PM and other methods should be reviewed from a PresPM perspective.

\bibliographystyle{myplainnat}

\footnotesize\bibliography{References}

\begin{thebibliography}{49}
\providecommand{\natexlab}[1]{#1}
\providecommand{\url}[1]{\texttt{#1}}
\expandafter\ifx\csname urlstyle\endcsname\relax
  \providecommand{\doi}[1]{doi: #1}\else
  \providecommand{\doi}{doi: \begingroup \urlstyle{rm}\Url}\fi

\bibitem[Bica et~al.(2020)Bica, Alaa, Jordon, and van~der Schaar]{PPM:seq2seq}
Bica, I., Alaa, A.~M., Jordon, J., and van~der Schaar, M.
\newblock Estimating counterfactual treatment outcomes over time through
  adversarially balanced representations.
\newblock \emph{CoRR}, abs/2002.04083, 2020.
\newblock URL \url{https://arxiv.org/abs/2002.04083}.

\bibitem[Bozorgi et~al.(2020)Bozorgi, Teinemaa, Dumas, La~Rosa, and
  Polyvyanyy]{PPM:meet}
Bozorgi, Z.~D., Teinemaa, I., Dumas, M., La~Rosa, M., and Polyvyanyy, A.
\newblock Process mining meets causal machine learning: Discovering causal
  rules from event logs.
\newblock In \emph{2020 2nd International Conference on Process Mining (ICPM)},
  pages 129--136, 2020.
\newblock \doi{10.1109/ICPM49681.2020.00028}.

\bibitem[Bozorgi et~al.(2021)Bozorgi, Teinemaa, Dumas, Rosa, and
  Polyvyanyy]{PPM:costaware}
Bozorgi, Z.~D., Teinemaa, I., Dumas, M., Rosa, M.~L., and Polyvyanyy, A.
\newblock Prescriptive process monitoring for cost-aware cycle time reduction.
\newblock In \emph{2021 3rd International Conference on Process Mining (ICPM)},
  pages 96--103, 2021.
\newblock \doi{10.1109/ICPM53251.2021.9576853}.

\bibitem[Bozorgi et~al.(2023)Bozorgi, Irene~Teinemaa, Dumas, La~Rosa, and
  Polyvyanyy]{PPM:ppmci}
Bozorgi, Z.~D., Irene~Teinemaa, I., Dumas, M., La~Rosa, M., and Polyvyanyy, A.
\newblock Prescriptive process monitoring based on causal effect estimation.
\newblock \emph{Information Systems}, 2023.
\newblock \doi{https://doi.org/10.1016/j.is.2023.102198}.
\newblock URL
  \url{https://www.sciencedirect.com/science/article/abs/pii/S0306437923000340}.

\bibitem[Branchi et~al.(2022)Branchi, Di~Francescomarino, Ghidini, Massimo,
  Ricci, and Ronzan]{PPM:learning}
Branchi, S., Di~Francescomarino, C., Ghidini, C., Massimo, D., Ricci, F., and
  Ronzan, M.
\newblock Learning to act: A reinforcement approach to learn to best
  activities.
\newblock In Di~Ciccio, C., Dijkman, R., del R{\'i}o~Ortega, A., and
  Rinderle-Ma, S., editors, \emph{Business Process Management Forum}, pages
  137--154, Cham, 2022. Springer International Publishing.
\newblock ISBN 978-3-031-16171-1.

\bibitem[Bugaenko(2021)]{PPM:replace}
Bugaenko, A.~A.
\newblock Application of reinforcement learning to optimize business processes
  in the bank.
\newblock \emph{Turkish Journal of Computer and Mathematics}, 12, 2021.

\bibitem[Burattin(2022)]{intro:stream_confcheck}
Burattin, A.
\newblock Streaming process mining.
\newblock In {van der Aalst}, W.~M. and Carmona, J., editors, \emph{Process
  Mining Handbook}, pages 349--372. Springer International Publishing, Cham,
  2022.
\newblock ISBN 978-3-031-08848-3.
\newblock \doi{10.1007/978-3-031-08848-3-1}.
\newblock URL \url{https://doi.org/10.1007/978-3-031-08848-3-1}.

\bibitem[Burggraef et~al.(2022)Burggraef, Steinberg, Heinbach, and
  Bamberg]{PPM:BURG}
Burggraef, P., Steinberg, F., Heinbach, B., and Bamberg, M.
\newblock Reinforcement learning for process time optimization in an assembly
  process utilizing an {Industry 4.0} demonstration cell.
\newblock \emph{Procedia CIRP}, 107:\penalty0 1095--1100, 2022.
\newblock \doi{https://doi.org/10.1016/j.procir.2022.05.114}.

\bibitem[Carvalho et~al.(2019)Carvalho, Soares, Vita, da~P.~Francisco, Basto,
  and Alcali\'{a}]{PPM:CARVAL}
Carvalho, T.~P., Soares, F.~A., Vita, R., da~P.~Francisco, R., Basto, J.~P.,
  and Alcali\'{a}, S.~G.
\newblock A systematic literature review of machine learning methods applied to
  predictive maintenance.
\newblock \emph{Computers \& Industrial Engineering}, 137:\penalty0 106024,
  2019.
\newblock ISSN 0360-8352.
\newblock \doi{https://doi.org/10.1016/j.cie.2019.106024}.

\bibitem[Dalzochio et~al.(2020)Dalzochio, Kunst, Pignaton, Binotto, Sanyal,
  Favilla, and Barbosa]{PPM:four}
Dalzochio, J., Kunst, R., Pignaton, E., Binotto, A., Sanyal, S., Favilla, J.,
  and Barbosa, J.
\newblock Machine learning and reasoning for predictive maintenance in
  {Industry 4.0}: Current status and challenges.
\newblock \emph{Computers in Industry}, 123:\penalty0 103298, 2020.
\newblock ISSN 0166-3615.
\newblock \doi{https://doi.org/10.1016/j.compind.2020.103298}.

\bibitem[{de Leoni} et~al.(2020){de Leoni}, Dees, and Reulink]{PPM:recommend}
{de Leoni}, M.~d., Dees, M., and Reulink, L.
\newblock Design and evaluation of a process-aware recommender system based on
  prescriptive analytics.
\newblock In \emph{2020 2nd International Conference on Process Mining (ICPM)},
  pages 9--16, 2020.
\newblock \doi{10.1109/ICPM49681.2020.00013}.

\bibitem[Devriendt et~al.(2021)Devriendt, Berrevoets, and Verbeke]{PPM:Wouter}
Devriendt, F., Berrevoets, J., and Verbeke, W.
\newblock Why you should stop predicting customer churn and start using uplift
  models.
\newblock \emph{Information Sciences}, 548:\penalty0 497--515, 2021.
\newblock ISSN 0020-0255.
\newblock \doi{https://doi.org/10.1016/j.ins.2019.12.075}.

\bibitem[Dumas et~al.(2022)Dumas, Fournier, Limonad, Marrella, Montali, Rehse,
  Accorsi, Calvanese, De~Giacomo, Fahland, et~al.]{PPM:manifesto}
Dumas, M., Fournier, F., Limonad, L., Marrella, A., Montali, M., Rehse, J.-R.,
  Accorsi, R., Calvanese, D., De~Giacomo, G., Fahland, D., et~al.
\newblock {AI}-augmented business process management systems: A research
  manifesto.
\newblock \emph{ACM Transactions on Management Information Systems}, 2022.

\bibitem[Fani~Sani et~al.(2023)Fani~Sani, Vazifehdoostirani, Park, Pegoraro,
  van Zelst, and {van der Aalst}]{concl:logsample}
Fani~Sani, M., Vazifehdoostirani, M., Park, G., Pegoraro, M., van Zelst, S.~J.,
  and {van der Aalst}, W.~M.
\newblock Performance-preserving event log sampling for predictive monitoring.
\newblock \emph{Journal of Intelligent Information Systems}, 2023.
\newblock ISSN 1573-7675.
\newblock \doi{10.1007/s10844-022-00775-9}.
\newblock URL \url{https://doi.org/10.1007/s10844-022-00775-9}.

\bibitem[Gottesman et~al.(2019)Gottesman, Johansson, Komorowski, Faisal,
  Sontag, Doshi-Velez, and Celi]{ppm:med}
Gottesman, O., Johansson, F., Komorowski, M., Faisal, A., Sontag, D.,
  Doshi-Velez, F., and Celi, L.~A.
\newblock Guidelines for reinforcement learning in healthcare.
\newblock \emph{Nature Medicine}, 25\penalty0 (1):\penalty0 16--18, jan 2019.
\newblock \doi{10.1038/s41591-018-0310-5}.
\newblock URL \url{https://doi.org/10.1038/s41591-018-0310-5}.

\bibitem[Huang et~al.(2011)Huang, {van der Aalst}, Lu, and Duan]{PPM:huang}
Huang, Z., {van der Aalst}, W.~M., Lu, X., and Duan, H.
\newblock Reinforcement learning based resource allocation in business process
  management.
\newblock \emph{Data \& Knowledge Engineering}, 70\penalty0 (1):\penalty0
  127--145, 2011.
\newblock ISSN 0169-023X.
\newblock \doi{https://doi.org/10.1016/j.datak.2010.09.002}.

\bibitem[Ibarz et~al.(2021)Ibarz, Tan, Finn, Kalakrishnan, Pastor, and
  Levine]{PPM:robotRL}
Ibarz, J., Tan, J., Finn, C., Kalakrishnan, M., Pastor, P., and Levine, S.
\newblock How to train your robot with deep reinforcement learning: Lessons we
  have learned.
\newblock \emph{The International Journal of Robotics Research}, 40\penalty0
  (4-5):\penalty0 698--721, 2021.
\newblock \doi{10.1177/0278364920987859}.
\newblock URL \url{https://doi.org/10.1177/0278364920987859}.

\bibitem[Imbens and Rubin(2015)]{PPM:imbens}
Imbens, G.~W. and Rubin, D.~B.
\newblock \emph{Causal Inference for Statistics, Social, and Biomedical
  Sciences: An Introduction}.
\newblock Cambridge University Press, 2015.
\newblock \doi{10.1017/CBO9781139025751}.

\bibitem[Koorn et~al.(2020)Koorn, Lu, Leopold, and Reijers]{intro:meaning}
Koorn, J.~J., Lu, X., Leopold, H., and Reijers, H.~A.
\newblock Looking for meaning: Discovering action-response-effect patterns
  in business processes.
\newblock In Fahland, D., Ghidini, C., Becker, J., and Dumas, M., editors,
  \emph{Business Process Management}, pages 167--183, Cham, 2020. Springer
  International Publishing.
\newblock ISBN 978-3-030-58666-9.

\bibitem[Kubrak et~al.(2021)Kubrak, Milani, Nolte, and Dumas]{PPM:quovadis}
Kubrak, K., Milani, F., Nolte, A., and Dumas, M.
\newblock Prescriptive process monitoring: Quo vadis?
\newblock \emph{CoRR}, abs/2112.01769, 2021.
\newblock URL \url{https://arxiv.org/abs/2112.01769}.

\bibitem[Leemans and Tax(2022)]{PPM:reason}
Leemans, S. J.~J. and Tax, N.
\newblock Causal reasoning over control-flow decisions in process models.
\newblock In Franch, X., Poels, G., Gailly, F., and Snoeck, M., editors,
  \emph{Adv. Information Systems Engineering}, pages 183--200, Cham, 2022.
  Springer International Publishing.
\newblock ISBN 978-3-031-07472-1.

\bibitem[Levine et~al.(2020)Levine, Kumar, Tucker, and Fu]{PPM:offline}
Levine, S., Kumar, A., Tucker, G., and Fu, J.
\newblock Offline reinforcement learning: Tutorial, review, and perspectives on
  open problems.
\newblock \emph{ArXiv}, abs/2005.01643, 2020.

\bibitem[Lisowska et~al.(2021)Lisowska, Wilk, and Peleg]{PPM:liso}
Lisowska, A., Wilk, S., and Peleg, M.
\newblock From personalized timely notification to healthy habit formation: A
  feasibility study of reinforcement learning approaches on synthetic data.
\newblock In \emph{SMARTERCARE@AI*IA}, pages 7--18, 2021.

\bibitem[Liu and Zou(2018)]{PPM:effects}
Liu, R. and Zou, J.
\newblock The effects of memory replay in reinforcement learning.
\newblock In \emph{2018 56th Annual Allerton Conference on Communication,
  Control, and Computing (Allerton)}, pages 478--485, 2018.
\newblock \doi{10.1109/ALLERTON.2018.8636075}.

\bibitem[Mnih et~al.(2015)Mnih, Kavukcuoglu, Silver, Veness, Bellemare, Graves,
  Riedmiller, Fidjeland, Ostrovski, Petersen, Beattie, Sadik, Antonoglou, King,
  Kumaran, Wierstra, Legg, and Hassabis]{PPM:replay}
Mnih, V., Kavukcuoglu, K., Silver, A.~A., David~Rusu, Veness, J., Bellemare,
  M.~G., Graves, A., Riedmiller, M., Fidjeland, A.~K., Ostrovski, G., Petersen,
  S., Beattie, C., Sadik, A., Antonoglou, I., King, H., Kumaran, D., Wierstra,
  D., Legg, S., and Hassabis, D.
\newblock Human-level control through deep reinforcement learning.
\newblock \emph{Nature}, 518:\penalty0 529--533, 2015.
\newblock \doi{10.1038/nature14236}.
\newblock URL \url{https://doi.org/10.1038/nature14236}.

\bibitem[Neal et~al.(2020)Neal, Huang, and Raghupathi]{PPM:Real}
Neal, B., Huang, C.-W., and Raghupathi, S.
\newblock Realcause: Realistic causal inference benchmarking, 2020.
\newblock URL \url{https://arxiv.org/abs/2011.15007}.

\bibitem[Olaya et~al.(2020{\natexlab{a}})Olaya, Coussement, and
  Verbeke]{PPM:Wouteruplift}
Olaya, D., Coussement, K., and Verbeke, W.
\newblock A survey and benchmarking study of multitreatment uplift modeling.
\newblock \emph{Data Mining and Knowledge Discovery}, 34:\penalty0 273--308,
  2020{\natexlab{a}}.
\newblock ISSN 1573-756X.
\newblock \doi{10.1007/s10618-019-00670-y}.
\newblock URL \url{https://doi.org/10.1007/s10618-019-00670-y}.

\bibitem[Olaya et~al.(2020{\natexlab{b}})Olaya, V{\'a}squez, Maldonado,
  Miranda, and Verbeke]{ppm:edu}
Olaya, D., V{\'a}squez, J., Maldonado, S., Miranda, J., and Verbeke, W.
\newblock Uplift modeling for preventing student dropout in higher education.
\newblock \emph{Decision support systems}, 134:\penalty0 113320,
  2020{\natexlab{b}}.

\bibitem[Paraschos et~al.(2020)Paraschos, Koulinas, and Koulouriotis]{PPM:PARA}
Paraschos, P.~D., Koulinas, G.~K., and Koulouriotis, D.~E.
\newblock Reinforcement learning for combined production-maintenance and
  quality control of a manufacturing system with deterioration failures.
\newblock \emph{Journal of Manufacturing Systems}, 56:\penalty0 470--483, 2020.
\newblock ISSN 0278-6125.
\newblock \doi{https://doi.org/10.1016/j.jmsy.2020.07.004}.

\bibitem[Park et~al.(2021)Park, Son, Ko, and Noh]{PPM:twins}
Park, K.~T., Son, Y.~H., Ko, S., and Noh, S.~D.
\newblock Digital twin and reinforcement learning-based resilient production
  control for micro smart factory.
\newblock \emph{Applied Sciences}, 11, 03 2021.
\newblock \doi{10.3390/app11072977}.

\bibitem[Pasquadibisceglie et~al.(2022)Pasquadibisceglie, Appice, Castellano,
  and Malerba]{PPM:cama}
Pasquadibisceglie, V., Appice, A., Castellano, G., and Malerba, D.
\newblock A multi-view deep learning approach for predictive business processes
  monitoring.
\newblock In \emph{2022 IEEE World Congress on Services (SERVICES)}, pages
  26--26, 2022.
\newblock \doi{10.1109/SERVICES55459.2022.00039}.

\bibitem[Pearl(2000)]{PPM:judea}
Pearl, J.
\newblock \emph{Causality: Models, Reasoning, and Inference}.
\newblock Cambridge University Press, Cambridge, 2000.
\newblock ISBN 0-521-77362-8.

\bibitem[Polyvyanyy et~al.(2019)Polyvyanyy, Pika, Wynn, and {ter
  Hofstede}]{intro:poly}
Polyvyanyy, A., Pika, A., Wynn, M.~T., and {ter Hofstede}, A.~H.
\newblock A systematic approach for discovering causal dependencies between
  observations and incidents in the health and safety domain.
\newblock \emph{Safety Science}, 118:\penalty0 345--354, 2019.
\newblock ISSN 0925-7535.
\newblock \doi{https://doi.org/10.1016/j.ssci.2019.04.045}.
\newblock URL
  \url{https://www.sciencedirect.com/science/article/pii/S0925753518316230}.

\bibitem[{Ponce de León} et~al.(2018){Ponce de León}, Nardelli, Carmona, and
  {vanden Broucke}]{PPM:neg}
{Ponce de León}, H., Nardelli, L., Carmona, J., and {vanden Broucke}, S.~K.
\newblock Incorporating negative information to process discovery of complex
  systems.
\newblock \emph{Information Sciences}, 422:\penalty0 480--496, 2018.
\newblock ISSN 0020-0255.
\newblock \doi{https://doi.org/10.1016/j.ins.2017.09.027}.
\newblock URL
  \url{https://www.sciencedirect.com/science/article/pii/S0020025516315444}.

\bibitem[Potoniec et~al.(2022)Potoniec, Sroka, and Pawlak]{PPM:discovercausal}
Potoniec, J., Sroka, D., and Pawlak, T.~P.
\newblock Continuous discovery of causal nets for non-stationary business
  processes using the online miner.
\newblock \emph{European Journal of Operational Research}, 303\penalty0
  (3):\penalty0 1304--1320, 2022.
\newblock ISSN 0377-2217.
\newblock \doi{https://doi.org/10.1016/j.ejor.2022.03.046}.
\newblock URL
  \url{https://www.sciencedirect.com/science/article/pii/S0377221722002685}.

\bibitem[Qafari and {van der Aalst}(2020)]{PPM:root}
Qafari, M.~S. and {van der Aalst}, W.~M.
\newblock Root cause analysis in process mining using structural equation
  models.
\newblock In Del R{\'i}o~Ortega, A., Leopold, H., and Santoro, F.~M., editors,
  \emph{Business Process Management Workshops}, pages 155--167, Cham, 2020.
  Springer International Publishing.
\newblock ISBN 978-3-030-66498-5.

\bibitem[Sallab et~al.(2017)Sallab, Abdou, Perot, and Yogamani]{ppm:RLcar}
Sallab, A.~E., Abdou, M., Perot, E., and Yogamani, S.
\newblock Deep reinforcement learning framework for autonomous driving.
\newblock \emph{Electronic Imaging}, 29\penalty0 (19):\penalty0 70--76, jan
  2017.
\newblock \doi{10.2352/issn.2470-1173.2017.19.avm-023}.
\newblock URL \url{https://doi.org/10.2352\%2Fissn.2470-1173.2017.19.avm-023}.

\bibitem[Scheibel and Rinderle-Ma(2022)]{conc:decmin}
Scheibel, B. and Rinderle-Ma, S.
\newblock Decision mining with time series data based on automatic feature
  generation.
\newblock In Franch, X., Poels, G., Gailly, F., and Snoeck, M., editors,
  \emph{Advanced Information Systems Engineering}, pages 3--18, Cham, 2022.
  Springer International Publishing.
\newblock ISBN 978-3-031-07472-1.

\bibitem[Sculley et~al.(2015)Sculley, Holt, Golovin, Davydov, Phillips, Ebner,
  Chaudhary, Young, Crespo, and Dennison]{PPM:hidden}
Sculley, D., Holt, G., Golovin, D., Davydov, E., Phillips, T., Ebner, D.,
  Chaudhary, V., Young, M., Crespo, J.-F., and Dennison, D.
\newblock Hidden technical debt in machine learning systems.
\newblock In Cortes, C., Lawrence, N., Lee, D., Sugiyama, M., and Garnett, R.,
  editors, \emph{Advances in Neural Information Processing Systems}, volume~28.
  Curran Associates, Inc., 2015.
\newblock URL
  \url{https://proceedings.neurips.cc/paper\_files/paper/2015/file/86df7dcfd896fcaf2674f757a2463eba-Paper.pdf}.

\bibitem[Shalit et~al.(2017)Shalit, Johansson, and Sontag]{PPM:ITE}
Shalit, U., Johansson, F.~D., and Sontag, D.
\newblock Estimating individual treatment effect: Generalization bounds and
  algorithms.
\newblock In Precup, D. and Teh, Y.~W., editors, \emph{Proceedings of the 34th
  International Conference on Machine Learning}, volume~70 of \emph{Proceedings
  of Machine Learning Research}, pages 3076--3085. PMLR, 06--11 Aug 2017.
\newblock URL \url{https://proceedings.mlr.press/v70/shalit17a.html}.

\bibitem[Shoush and Dumas(2022)]{PPM:when}
Shoush, M. and Dumas, M.
\newblock When to intervene? prescriptive process monitoring under uncertainty
  and resource constraints.
\newblock In Di~Ciccio, C., Dijkman, R., del R{\'i}o~Ortega, A., and
  Rinderle-Ma, S., editors, \emph{Business Process Management Forum}, pages
  207--223, Cham, 2022. Springer International Publishing.
\newblock ISBN 978-3-031-16171-1.

\bibitem[Spielberg et~al.(2020)Spielberg, Tulsyan, Lawrence, Loewen, and
  Gopaluni]{PPM:Spielberg}
Spielberg, S.~P., Tulsyan, A., Lawrence, N.~P., Loewen, P.~D., and Gopaluni,
  R.~B.
\newblock Deep reinforcement learning for process control: A primer for
  beginners.
\newblock \emph{ArXiv}, abs/2004.05490, 2020.

\bibitem[Sutton and Barto(2018)]{PPM:RL}
Sutton, R.~S. and Barto, A.~G.
\newblock \emph{Reinforcement Learning: An Introduction}.
\newblock MIT Press, Cambridge, 2018.
\newblock ISBN 9780262039246.

\bibitem[Tan et~al.(2018)Tan, Zhang, Coumans, Iscen, Bai, Hafner, Bohez, and
  Vanhoucke]{PPM:quadsim}
Tan, J., Zhang, T., Coumans, E., Iscen, A., Bai, Y., Hafner, D., Bohez, S., and
  Vanhoucke, V.
\newblock Sim-to-real: Learning agile locomotion for quadruped robots.
\newblock \emph{arXiv preprint arXiv:1804.10332}, 2018.

\bibitem[{van der Aalst} et~al.(2021){van der Aalst}, Hinz, and
  Weinhardt]{conc:digtwin}
{van der Aalst}, W.~M., Hinz, O., and Weinhardt, C.
\newblock Resilient digital twins.
\newblock \emph{Business \& Information Systems Engineering}, 63:\penalty0
  615--619, 2021.
\newblock ISSN 1867-0202.
\newblock \doi{10.1007/s12599-021-00721-z}.
\newblock URL \url{https://doi.org/10.1007/s12599-021-00721-z}.

\bibitem[Weinzierl et~al.(2020)Weinzierl, Dunzer, Zilker, and
  Matzner]{PPM:nextbest}
Weinzierl, S., Dunzer, S., Zilker, S., and Matzner, M.
\newblock Prescriptive business process monitoring for recommending next best
  actions.
\newblock In Fahland, D., Ghidini, C., Becker, J., and Dumas, M., editors,
  \emph{Business Process Management Forum}, pages 193--209, Cham, 2020.
  Springer International Publishing.
\newblock ISBN 978-3-030-58638-6.

\bibitem[Weytjens and {De Weerdt}(2022)]{PPM:unc}
Weytjens, H. and {De Weerdt}, J.
\newblock Learning uncertainty with artificial neural networks for predictive
  process monitoring.
\newblock \emph{Applied Soft Computing}, 125:\penalty0 109134, 2022.
\newblock ISSN 1568-4946.
\newblock \doi{https://doi.org/10.1016/j.asoc.2022.109134}.

\bibitem[Winston(2022)]{ppm:ORbook}
Winston, W.~L.
\newblock \emph{Operations research: applications and algorithms}.
\newblock Cengage Learning, 2022.

\bibitem[Zeng et~al.(2023)Zeng, Cai, Sun, Huang, and Hao]{PPM:CRL}
Zeng, Y., Cai, R., Sun, F., Huang, L., and Hao, Z.
\newblock A survey on causal reinforcement learning, 2023.
\newblock URL \url{https://arxiv.org/abs/2302.05209}.

\end{thebibliography}
\end{document}